\documentclass[letterpaper]{article} 
\usepackage{aaai2026}  
\usepackage{times}  
\usepackage{helvet}  
\usepackage{courier}  
\usepackage[hyphens]{url}  
\usepackage{graphicx} 
\usepackage{natbib}  
\usepackage{caption} 
\usepackage{algorithm}
\usepackage{algorithmic}

\usepackage{amsfonts} 
\usepackage{mathtools} 
\usepackage{amsthm} 
\usepackage{bm} 
\usepackage{multirow} 
\usepackage{makecell} 
\usepackage{booktabs} 
\usepackage{amsmath} 
\usepackage{subfig} 
\usepackage{graphicx} 
\usepackage{amssymb} 
\usepackage{pifont} 
\usepackage{algorithm} 
\usepackage{algorithmic} 
\usepackage{xcolor} 
\usepackage{colortbl} 
\usepackage{bbding} 

\usepackage{newfloat}
\usepackage{listings}
\DeclareCaptionStyle{ruled}{labelfont=normalfont,labelsep=colon,strut=off} 
\floatstyle{ruled}
\newfloat{listing}{tb}{lst}{}
\floatname{listing}{Listing}
\title{MP-ISMoE: Mixed-Precision Interactive Side Mixture-of-Experts\\ for Efficient Transfer Learning}
\author{
    Yutong Zhang\textsuperscript{\rm 1,\rm 2},
    Zimeng Wu\textsuperscript{\rm 1,\rm 2},
    Shengcai Liao\textsuperscript{\rm 3},
    Shujiang Wu\textsuperscript{\rm 2},
    Jiaxin Chen\textsuperscript{\rm 1,\rm 2,}\thanks{Corresponding Author}
}
\affiliations{
    \textsuperscript{\rm 1}State Key Laboratory of Virtual Reality Technology and Systems, Beihang University, Beijing, China\\
    \textsuperscript{\rm 2}School of Computer Science and Engineering, Beihang University, Beijing, China\\
    \textsuperscript{\rm 3}College of Information Technology, United Arab Emirates University, Al Ain, Abu Dhabi, United Arab Emirates\\
}

\usepackage{bibentry}

\begin{document}

\maketitle

\begin{abstract}
Parameter-efficient transfer learning (PETL) has emerged as a pivotal paradigm for adapting pre-trained foundation models to downstream tasks, significantly reducing trainable parameters yet suffering from substantial memory overhead caused by gradient backpropagation during fine-tuning. While memory-efficient transfer learning (METL) circumvents this challenge by bypassing backbone gradient computation via lightweight small side networks, its stringent memory constraint severely limits learning capacity of side networks, thereby significantly compromising performance. To address these limitations, we propose a novel Mixed-Precision Interactive Side Mixture-of-Experts framework (MP-ISMoE). Specifically, we first propose a Gaussian Noise Perturbed Iterative Quantization (GNP-IQ) scheme to quantize weights into lower-bits while effectively decreasing quantization errors. By leveraging memory conserved from GNP-IQ, we subsequently employ Interactive Side Mixture-of-Experts (ISMoE) to scaling up side networks without sacrificing overall memory efficiency. Different from conventional mixture-of-experts, ISMoE learns to select optimal experts by interacting with salient features from frozen backbones, thus suppressing knowledge forgetting and boosting performance. Extensive experiments across diverse vision-language and language-only tasks demonstrate that MP-ISMoE remarkably promotes accuracy compared to state-of-the-art METL approaches, while maintaining comparable parameter and memory efficiency. 
\end{abstract}

\begin{links}
    \link{Code \& Extended version}{https://github.com/Zhang-VKk/MP-ISMoE.git}
\end{links}

\section{Introduction}
\label{introduction}
Large-scale foundation models, which are typically pre-trained on massive datasets, have demonstrated remarkable representation and generalization abilities across a wide range of domains, including computer vision \cite{fang2023eva}, natural language processing \cite{touvron2023llama}, and multi-modal tasks \cite{wu2024mixture}. 
Transfer learning effectively unleashes the potential of these models, facilitating development of numerous task-specific models.
However, with the ever-expanding model scale, the fully fine-tuning paradigm \cite{lv2023full,yin20255} becomes prohibitively resource-intensive.

Parameter-Efficient Transfer Learning (PETL) \cite{houlsby2019param,li2021prefix,hu2022lora} has become a promising solution in balancing training cost and model capacity, by freezing most parameters of backbones and fine-tuning lightweight modules such as partial parameters \cite{zaken2022bitfit}, prompt vectors \cite{li2021prefix} or adapter networks \cite{chen2022adaptformer}.
However, they still require backpropagation through large backbones, leading to excessive memory consumption that is disproportionate to the reduction in trainable parameters.

More recently, Memory-Efficient Transfer Learning (METL) \cite{zhang2020side,sung2022lst,liu2024tuning,mercea2024time} has emerged to achieve consistent reduction in both trainable parameters and memory overhead. Typically, METL introduces a lightweight trainable side network parallel to the frozen backbone, connecting paired features via ladder modules. It primarily turns gradient backpropagation to tiny side networks and  ladders, thus improving training efficiency. However, existing METL methods predominantly focus on improving the efficiency and representational capacity of side networks, suffering from the following issues. 
(1) \emph{Sub-optimal allocation of memory budget.} Given the dominant memory footprint of backbones, existing methods often allocate a small portion of memory to the side network, along with a stringent constraint on the amount of parameters. It inherently restrains the learning capacity of the side network, thus substantially limiting ultimate performance.
(2) \emph{Rigid and simplistic side network structure.} The side network is typically down-scaled proportionally from the backbone, inheriting its dense activation feature. Such fixed design leads to a sub-optimal trade-off between memory efficiency and model capacity, hampering effective transfer learning under tight memory budgets.
(3) \emph{Insufficient exploration of guidance from backbones.} Most existing METL methods leverage features from backbones to fine-tune side networks by directly combining it with those from side networks through weighted summation. They fail to elaborately explore the complementary knowledge from backbone to suppress over-fitting and knowledge forgetting, thus leaving much room for improvement.

To address above limitations, we propose a novel METL method, dubbed Mixed-Precision Interactive Side Mixture-of-Experts (\textbf{MP-ISMoE}). 
MP-ISMoE aims at enabling more effective memory allocation between backbone and side network, while facilitating efficient side network expansion during fine-tuning. To this end, we first introduce Gaussian Noise Perturbed Iterative Quantization (\textbf{GNP-IQ}), which applies iterative weight quantization with Gaussian noise perturbation to the backbone network, reducing massive memory footprint while maintaining model performance. Leveraging saved memory space by GNP-IQ, we design the Interactive Side Mixture-of-Experts (\textbf{ISMoE}) to expand the capacity of side network by employing a sparse MoE structure, without violating the overall memory budget. ISMoE further explores class tokens from backbone as guidance of general knowledge, and interactively apply them to adjust expert selection based on their correlations, thereby mitigating knowledge forgetting and over-fitting.

In summary, our main contributions lie in three-fold:

\begin{itemize}
    \item We propose a novel mixed-precision fine-tuning framework with an MoE-based side network, dubbed as MP-ISMoE, for memory efficient transfer learning.
    \item We propose a Gaussian Noise Perturbed Iterative Quantization (GNP-IQ) process and an Interactive Side Mixture-of-Experts (ISMoE) structure to optimize memory-constrained resource allocation, expand the capacity of side network and mitigate knowledge forgetting, thereby enhancing transfer learning performance.
    \item We conduct extensive experiments on both vision-language and natural language processing tasks across multiple network architectures, demonstrating that MP-ISMoE remarkably outperforms the state-of-the-art METL methods, with comparable parameter and memory efficiency.
\end{itemize}

\section{Related Work}
\label{related_work}
\subsection{Parameter-Efficient Transfer Learning}
Current PETL methods can be categorized into three paradigms: (1) $\textit{Partial Tuning}$ updates only a subset of original parameters, such bias \cite{zaken2022bitfit}, weights \cite{touvron2022three}, and task-specific ones \cite{sung2021training},  while keeping the rest frozen. (2) $\textit{Prompt Tuning}$ embeds sparse manual-tuning \cite{zhang2023multimodal} or dense randomly-initialized \cite{zhou2022conditional} tokens into input or intermediate state of the backbone. (3) $\textit{Adapter Tuning}$ introduces lightweight learnable modules into frozen backbones, some methods \cite{hu2022lora,liu2022few,mao2025survey} employ scaling/shifting factors, learnable vectors, or compact MLPs to refine feature projections, while others \cite{jie2023fact,zhang2023adalora} minimize trainable parameters via matrix decomposition and hyper-network prediction. 

\subsection{Memory-Efficient Transfer Learning}
The METL endeavors to achieve the optimal performance-memory balance in resource-constrained scenarios. 
(1) $\textit{General Memory Optimization}$. 
Mixed-precision training \cite{micikevicius2017mixed} or quantization strategies \cite{wang2018training} introduce low bitwidth formats for weights, activations and gradients.
Gradient checkpoint \cite{2016Training} selectively store critical intermediates, reconstructing \cite{gomez2017reversible} discarded activations during backpropagation. 
(2) $\textit{Backpropagation Decoupling}$. Orthogonally, another direction isolates gradient computation for extensive parameters of large models. Some manners \cite{raffel2020exploring} update an extra projection layer, which follows the last backbone layer. While other methods \cite{zhang2020side,sung2022lst,diao2024unipt,diao2024sherl} introduce a parallel lightweight network to augment the static main network for new domains.

\begin{figure*}[!ht]
\centering
\includegraphics[width=1\textwidth]{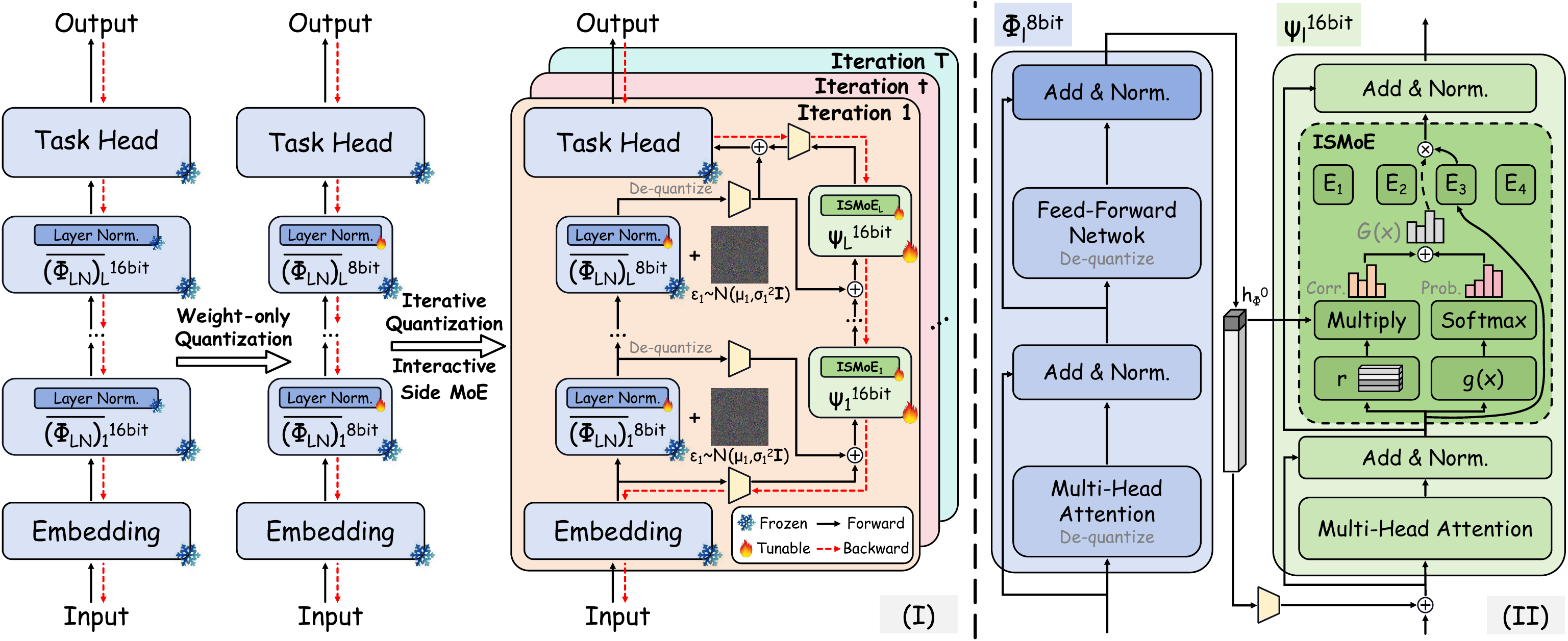} 
\caption{Diagram on the proposed framework. \textbf{Part (I)} presents Mixed-Precision Interactive Side Mixture-of-Experts for universal METL based on side networks. We first quantize most of weights in pre-trained backbone before fine-tuning, while preserving trainable Layer Normalization with full-precision. Subsequently, we introduce a learnable parallel Interactive Side Mixture-of-Experts (ISMoE) structure, connected to backbone via downsampling module. During fine-tuning phase, Gaussian Noise Perturbed Iterative Quantization (GNP-IQ) strategy introduces
Gaussian noise perturbations into weights to mitigate quantization error. 
\textbf{Part (II)} illustrates detailed structure of $l^{th}$ layer of MP-ISMoE. We adopt MoE to scale up side network, design representative feature for each expert to measure correlation with backbone, and select Topk experts based on the routing Probability and Correlation Score.}
\label{fig-framework}
\end{figure*}

\subsection{Weight-only Post-Training Quantization} 
The Weight-only Post-Training Quantization \cite{frantar2022gptq,kim2023squeezellm,lee2023owq,lin2024awq} proves effective in accelerating the memory-bounded General Matrix-Vector Multiply (GEMV) operators while aims to convert weights from high-precision to low-precision with fewer bits, thus reducing the size of model and speeding up weight loading. 

\subsection{Mixture-of-Experts}
Mixture-of-Experts (MoE) \cite{cai2025survey,mu2025comprehensive} divides a model into specialized components (\emph{i.e.}, Experts), each of which handles distinct tasks or data aspects, and combines the router \cite{liu2024routers,harvey2025optimizing} to selectively activate relevant experts, thereby leveraging a vast amount of expertise by increasing model capacity while maintaining computational efficiency. Typically, MoE can be categorized into two variants: \textit{Dense MoE} \cite{pan2024dense,wumixture} activates all experts in each iteration, while \textit{Sparse MoE} \cite{dai2024deepseekmoe,lieber2024jamba,wei2024skywork} activates only some experts and thus generally has lower computational overhead.

\section{Methodology}

\subsection{Framework Overview}
Existing METL approaches based on side network typically impose strict constraints on the scale of trainable parameters, aiming to ensure low memory overhead during training. 
However, such limitation tend to compromise the representation capacity of model, thus leading to sub-optimal performance on downstream tasks. 
Therefore, we propose a novel framework, Mixed-Precision Interactive Side Mixture-of-Experts (MP-ISMoE), as depicted in Figure~\ref{fig-framework}. 

Specifically, Gaussian Noise Perturbed Iterative Quantization (GNP-IQ) module saves memory consumption of backbone weights by strategically reducing their numerical precision. To mitigate the increasing quantization error accumulated during fine-tuning, GNP-IQ employs an iterative re-quantization with injected Gaussian noise perturbation. Meanwhile, the Interactive Side Mixture-of-Experts (ISMoE) module improves the scalability of the side branch by introducing a memory-efficient MoE-based structure. To address the catastrophic knowledge forgetting issue, a cross-network representative token interaction mechanism is performed.

As for the overall fine-tuning process, the majority of the backbone parameters are frozen in 8-bit precision, while the Layer Normalization parameters remain trainable in full precision, and the side network is trained in 16-bit. In summary, our MP-ISMoE framework enables efficient mixed-precision fine-tuning, thereby achieving a superior trade-off between memory efficiency and downstream performance.

\subsection{Gaussian Noise Perturbed Iterative Quantization}
\label{GNP-IQ}
\paragraph{Quantization of Backbone.} 
In context of a Transformer-based backbone network, we denote the set of parameters belonging to all Layer Normalization layers as $\Phi_{\text{LN}}$, and the rest of the parameters as $\overline{\Phi_{\text{LN}}}$. Typically for memory efficient fine-tuning, $\Phi_{\text{LN}}$ and $\overline{\Phi_{\text{LN}}}$ are kept frozen.
To further reduce the memory footprint of the backbone network, we perform weight-only quantization on $\overline{\Phi_{\text{LN}}}$.

Specifically, prior to fine-tuning, an initial asymmetric post-training quantization is applied, converting the full-precision weights into compact low-bit ones, formally as:
\begin{equation}\label{quantization}
    w_q=\textrm{clamp}(\lfloor{\frac{w_f}{s}}\rfloor+z;0;2^n-1),
\end{equation}
where $w_f \in \overline{\Phi_{\text{LN}}}$ denotes the original floating-point weight, and $w_q$ is its quantized fixed-point counterpart. Here $n$ is the bitwidth of the quantized value, and $\textrm{clamp}(\cdot;a;b)$ denotes truncating the value to the interval $[a,b]$. The scale factor $s$ and zero-point $z$ are calculated as:
\begin{equation}\label{quatization_sz}
    \begin{aligned}
        s&=\frac{r_{\max}-r_{\min}}{q_{\max}-q_{\min}}=\frac{r_{\max}-r_{\min}}{2^n-1},\\
        z&=\textrm{clamp}(\lfloor q_{\max}-\frac{r_{\max}}{s}\rfloor;0;2^n-1),
    \end{aligned}
\end{equation}
where $r_{{\min}/{\max}}$ and $q_{{\min}/{\max}}$ represent the numerical range of $w_f$ and $w_q$, respectively. While $w_q$ is stored in low-bit format, it is dequantized back to full-precision during forward pass computation by reversing Eq.~\eqref{quatization_sz} as:
\begin{equation}\label{de-quantization}
    w_d=s\cdot (w_q-z).
\end{equation}

\paragraph{Iterative Quantization Strategy.}
Although most of the backbone parameters are frozen during fine-tuning, the aforementioned initially assigned quantization coefficients cannot remain fixed throughout the entire process.
Noting that a small portion of quantized weights are updated during fine-tuning, the originally determined $s$ and $z$ may no longer be optimal for the evolving weight distribution. This mismatch leads to an increased quantization error $\text{Error}_{q}$, which can be expressed as:
\begin{equation}\label{quantization_error}
    \text{Error}_{q}=\frac{1}{U}\sum_{u=1}^{U}{(w_f^{(u)}-w_d^{(u)})}^2,
\end{equation}
where $U=|\overline{\Phi_{\text{LN}}}|$ denotes the number of quantized weights.

To mitigate this error, we introduce an iterative re-quantization mechanism for the backbone. To be precise, each time we randomly sample a small fraction $p\%$ of weights from $\overline{\Phi_{\text{LN}}}$ and re-compute their scale and zero-point coefficients (\emph{i.e.} $s$ and $z$), thereby refining the quantization to better fit the evolving state of the model. 
To save the additional cost for quantization, this step is performed iteratively at a fixed interval of $M$ epochs. Overall, this procedure is conducted $T=N_{\text{epoch}}/{M}$ times during fine-tuning, where $N_{\text{epoch}}$ is the total number of fine-tuning epochs.

\paragraph{Gaussian Noise Perturbation.}
Furthermore, to bridge the gap between infrequent quantization steps and the continue shift in model dynamics, we inject a Gaussian noise perturbation into weights prior to each re-quantization, formally as Eq.~(\ref{add_noise}). 
\begin{equation}\label{add_noise}
    w_f' = w_f+\epsilon_t,~\text{when}~n_{\text{epoch}}=t\times M.
\end{equation}
Here, $w_f'$ denotes the perturbed weight for further quantization, and $n_{\text{epoch}}$ is the index of current epoch. $\epsilon_t \sim \mathcal{N}(\mu_t,\sigma_t^2\mathbf{I}),t\in1,2,\cdots,T$ is a learnable perturbation draw from a Gaussian distribution. By optimizing the mean $\mu_t$ and standard variance $\sigma_t$, the perturbation serves to simulate the cumulative effect of latent parameter updates between re-quantizations, allowing the backbone to better anticipate its optimal quantized state during fine-tuning.

\subsection{Interactive Side Mixture-of-Experts}
\label{ISMoE}
\paragraph{Sparse MoE Based Side Network.}
With GNP-IQ scheme enabling a memory efficient backbone, we allocate the saved resources to expand the side network, thus enhancing its representation capability, while maintaining the overall memory budget.
However, a direct dense expansion on the width of network falls short of achieving an optimal performance-efficiency balance, owing to the substantial increase in trainable parameters and memory consumption.
Therefore, we adopt a sparse MoE in side network, enabling substantial capacity expansion under limiter memory constraints.

Concretely, ISMoE constructs a set of $N$ distinct experts $\{E_i\}^N_{i=1}$ by replicating the original FFN blocks from the side network. A sparse gating mechanism is then introduced to dynamically select the top-$k$ most relevant experts for each input token.
To be precise, a linear projection function $g(\cdot)\in \mathbb{R}^N$ first computes raw gating scores over experts. These scores are then sparsified using a masked top-$k$ selection operator $\mathcal{M}_k(\cdot)$, which retains only the top-$k$ values and set the rest to $-\infty$. The resulting sparse scores are normalized via a Softmax function to produce the final routing probability ${G}(\cdot)$. Let $\bm{x}^{in}$ and $\bm{x}^{out}$ denote the input and output feature of the ISMoE module, respectively, the final output is computed as the probability weighted sum of the selected expert outputs. The complete forward pass computation is formally defined as Eq.~\eqref{moe_gate}.
\begin{equation}\label{moe_gate}
    \begin{aligned}
        {G}(\bm{x})&=\textrm{Softmax}(\mathcal{M}_k(g(\bm{x}))),\\
        \mathcal{M}_k(\bm{x})_i&=
        \begin{cases}
            \bm{x}_i, \enspace \text{if}~\bm{x}_i\in \text{top-}k(\bm{x})).\\
            -\infty, \enspace \text{otherwise},
        \end{cases} \\
        \bm{x}^{out} &= \sum_{i=1}^{N}{G}_i(\bm{x}^{in})\cdot E_i(\bm{x}^{in}).
    \end{aligned}
\end{equation}

\paragraph{Cross-Network Interaction Guided Expert Selection.}
Although the MoE-based side network benefits from flexibly activated experts, unconstrained fully training of these parameters may entangle the roles of side experts and the backbone, which causes over-fitting to task-specific patterns and forgetting of general knowledge, ultimately leading to sub-optimal performance.
To address this issue, we introduce a cross-network interaction mechanism that explicitly serializes the capability of both branches, thereby fostering the learning of complementary knowledge across different network branches.
In particular, the general knowledge encoded in the backbone is directly utilized to guide expert selection in the side network, ensuring complementary collaboration between two branches.

Specifically, we extract a salient token ${\bm{h}_\Phi}^0\in\mathbb{R}^D$, which typically refers to the [CLS] token in context of the Transformer-based architecture, from the backbone as a proxy for general-purpose representations.
For experts in the side network, we initiate a learnable matrix of representative tokens $\bm{r}\in \mathbb{R}^{N\times D}$, where each row $\bm{r}_i$ corresponds to the affinity of expert $E_i$ with the requirement from general knowledge.
Then, a correlation score vector $\bm{c} \in \mathbb{R}^{N}$ is computed by measuring the similarity between the similarity between the salient token and the expert-wise representative tokens as Eq.~\eqref{similarity}, which serve as a general-knowledge-informed prior over the expert selection process.
\begin{equation}\label{similarity}
    \bm{c}=\textrm{Norm}({\bm{h}_\Phi}^0\times \bm{r}),
\end{equation}
where $\textrm{Norm}(\cdot)$ represents normalization operation. 
Finally, we integrate this prior into the expert selection operation by modulating the routing probability for top-$k$ expert selection:
\begin{equation}\label{final_selection}
    g'(\bm{x}) = \frac{\textrm{Softmax}(g(\bm{x}))+c}{2},
\end{equation}
where $g'(\cdot)$ denotes the refined routing probability, and the subsequent gating operations follow the same process as Eq.~(\ref{moe_gate}).

\begin{table*}[!t]
    \begin{center}
    \setlength{\tabcolsep}{0.36pt}
    \resizebox{\textwidth}{!}{
    \begin{tabular}{l c c ccc ccc ccc | c c ccc ccc}
    \toprule[1.2pt]
    
    \multirow{2}{*}{Method} & Params. & \multicolumn{1}{c}{Mem.} & \multicolumn{3}{c}{Flickr30K} & \multicolumn{3}{c}{MSCOCO1K} & \multicolumn{3}{c|}{MSCOCO5K} & Params. & \multicolumn{1}{c}{Mem.} & \multicolumn{3}{c}{MSR-VTT} & \multicolumn{3}{c}{MSVD} \\
    
    \cmidrule[0.4pt]{4-12}
    \cmidrule[0.4pt]{15-20}
    & (M) $\!\downarrow$ & (G) $\!\downarrow$ & I-T $\!\uparrow\ $ & T-I $\!\uparrow\ $ & Rsum $\!\uparrow\ $ & I-T $\!\uparrow\ $ & T-I $\!\uparrow\ $ & Rsum $\!\uparrow\ $ & I-T $\!\uparrow\ $ & T-I $\!\uparrow\ $ & Rsum $\!\uparrow\ $ & (M) $\!\downarrow$ &  (G) $\!\downarrow$ & T-V $\!\uparrow\ $ & V-T $\!\uparrow\ $ & Rsum $\!\uparrow\ $ & T-V $\!\uparrow\ $ & V-T $\!\uparrow\ $ & Rsum $\!\uparrow\ $ \\
    
    \midrule[0.6pt]
    {Fully-FT} &  {201.2} &  {176.8} &  {85.6} &  {73.3} &  {546.6} &  {83.1} &  {71.7} &  {542.7} &  {64.2} &  {51.2} &  {468.9} &  {151.3} &  {48.8} &  {42.8} &  {42.1} &  {389.2} &  {45.2} &  {57.1} &  {425.5}  \\
    \midrule[0.6pt]
    LST & \textbf{9.7} & \textbf{24.4} & 82.1 & 66.5 & 529.5 & 78.2 & 64.8 & 525.8 & 57.8 & 43.1 & 434.5 & 11.2 & 32.0 & 37.0 & 37.8 & 356.7 & 35.5 & 55.4 & 407.2\\
    UniPT & 12.4 & \textbf{24.4} & 84.8 & 69.1 & 537.4 & 80.6 & 67.5 & 532.9 & 61.1 & 45.9 & 445.3 & \textbf{9.6} & \textbf{13.6} & 38.9 & 39.3 & 361.3 & 40.9 & 59.7 & 432.1\\
    SHERL & \underline{11.3} & \textbf{24.4} & 86.1 & {71.1} & {542.3} & {81.8} & \underline{69.2} & {537.5} & \underline{62.5} & \underline{47.3} & \underline{450.8} & \textbf{9.6} & \textbf{13.6} & 39.2 & 40.6 & 363.7 & 40.9 & {60.2} & 429.7\\
   
    \textbf{Ours${}^\dagger$} & 12.9 & {25.5} & \underline{86.5} & \underline{71.3} & \underline{543.1} & \underline{81.9} & {69.1} & \underline{538.7} & 62.2 & 47.1 & 449.1 & \underline{10.1} & \underline{14.5} & \underline{39.9} & \underline{41.1} & \underline{365.5} & \underline{42.0} & \underline{60.4} & \underline{435.6}\\
    
    \textbf{Ours${}^\ddagger$} & 11.8 & \underline{25.4} & \textbf{87.4} & \textbf{73.3} & \textbf{547.0} & \textbf{82.8} & \textbf{71.0} & \textbf{542.6} & \textbf{63.4} & \textbf{48.7} & \textbf{453.8}& \underline{10.1} & {14.6} & \textbf{40.3} & \textbf{41.3} & \textbf{366.9} & \textbf{42.1} & \textbf{60.7} & \textbf{435.8}\\
    \bottomrule[1.2pt]
    \end{tabular}
    }
    \resizebox{\textwidth}{!}{
    \begin{tabular}{l c c cc cc | c c c ccc ccc cc}
    \multirow{2}{*}{Method} & Params. & \multicolumn{1}{c}{Mem.} & \multicolumn{2}{c}{VQAv2} & \multicolumn{2}{c|}{GQA} 
    & Params. & \multicolumn{1}{c}{Mem.} & \multicolumn{3}{c}{RefCOCO} & \multicolumn{3}{c}{RefCOCO+} & \multicolumn{2}{c}{RefCOCOg}\\
    
    \cmidrule[0.4pt]{4-7}
    \cmidrule[0.4pt]{10-17}
    & (M) $\!\downarrow$ & (G) $\!\downarrow$ & Test$_\text{D}$ $\!\uparrow$ & Test$_\text{S}$ $\!\uparrow$ & Test$_\text{D}$ $\!\uparrow$ & Test$_\text{S}$ $\!\uparrow\ $ & (M) $\!\downarrow$ & (G) $\!\downarrow$ & Val $\!\uparrow$ & TestA $\!\uparrow$ & TestB $\!\uparrow$ & Val $\!\uparrow$ & TestA $\!\uparrow$ & TestB $\!\uparrow$ & Val $\!\uparrow$ & Test $\!\uparrow$\\
    
    \midrule[0.6pt]
     {Fully-FT} &  {236.8} &  {82.0} & {76.71} &  {76.86} &  {60.25} &  {61.44}&  {185.2} &  {39.6} & {86.51} &  {89.13} &  {81.22} &  {79.54} &  {84.54} &  {70.63} &  {80.92} &  {80.95} \\
    \midrule[0.6pt]
    LST & 13.4 & 25.6 & 75.29 & 75.44 & 59.93 & 60.75 & 0.9 & 12.6 & 81.63 & 85.19 & 76.03 & 71.32 & 78.20 & 62.06 & 72.53 & 73.67\\
    UniPT & \textbf{10.3} & \textbf{11.6} & 75.33 & 75.53 & 60.10 & 60.72 & \textbf{0.7} & \textbf{6.8} & 82.71 & 86.25 & 78.16 & 72.94 & 79.18 & 64.49 & 77.04 & 77.33\\
    SHERL & 13.0 & 14.0 & 75.53 & 75.82 & 60.16 & 60.82 & \textbf{0.7} & \textbf{6.8} & 83.02 & 86.39 & 78.41 & 73.29 & \underline{80.11} & 64.59 & {77.80} & 77.33\\
    
    \textbf{Ours${}^\dagger$} & \underline{10.9} & \underline{12.6} & \underline{75.82} & \underline{76.87} & \underline{60.78} & \underline{61.41} & \underline{0.8} & \underline{7.3} & \underline{83.32} & \underline{87.09} & \textbf{79.20} & \underline{73.59} & {79.68} & \underline{64.88} & \underline{77.86} & \underline{77.95}\\
    
    \textbf{Ours${}^\ddagger$} & 13.6 & 14.9 & \textbf{76.21} & \textbf{76.91} & \textbf{60.91} & \textbf{61.44} & \underline{0.8} & {7.4} & \textbf{83.49} & \textbf{87.26} & \underline{79.19} & \textbf{73.92} & \textbf{80.51} & \textbf{65.02} &\textbf{78.39} & \textbf{78.08}\\
    \bottomrule[1.2pt]
    \end{tabular}}
    \end{center}
    \caption{Comparison results (\%) with METL approaches across various architectures and distinct VL tasks, in terms of amount of learnable parameters, memory usage, and other task-specific metrics. The best results are highlighted in \textbf{bold}, and the second best results are \underline{underlined}.}    \label{VL_table}
\end{table*}

\section{Experimental Results and Analysis}
\label{experiment}

\subsection{Experimental Settings}
\paragraph{Datasets and Evaluation Metrics.} 
We validate our proposed MP-ISMoE on both Vision-Language (VL) and Natural Language Processing (NLP) tasks. Specifically, for VL tasks, we conduct experiments on \textit{image-text retrieval} (ITR: Flickr30K \cite{young2014image}, MSCOCO \cite{lin2014microsoft}), \textit{video-text retrieval} (VTR: MSVD \cite{chen2011collecting}, MSR-VTT \cite{xu2016msr}), \textit{visual and compositional question answering} (VQA: VQAv2 \cite{goyal2017making}, GQA: GQA \cite{hudson2019gqa}), and \textit{visual grounding} (VG: RefCOCO, RefCOCO+ \cite{yu2016modeling}, RefCOCOg \cite{mao2016generation}). 
By following \cite{diao2024unipt}, we report Recall@1 (R@1) and Rsum of R@1,5,10 on cross-modal retrieval tasks, overall Accuracy on QA tasks, and mean Average Precision (mAP) on VG tasks. In the case of NLP task, we adopt GLUE benchmark \cite{wang2018glue} and present Accuracy Metric, F1 Score, Matthew's Correlation, Pearson-Spearman Correlation as the evaluation metrics for various datasets respectively. Detailed descriptions are provided in \textit{Extended Version}.

\paragraph{Counterparts.} We compare MP-ISMoE with full fine-tuning and two representative efficient adaptation paradigms: (1) \textit{Memory-Efficient} approaches exemplified by LST \cite{sung2022lst}, UniPT \cite{diao2024unipt}, and SHERL \cite{diao2024sherl}; (2) \textit{Parameter-Efficient} methods including Partial Tuning (BitFit \cite{zaken2022bitfit}), Prompt Tuning (Prompt \cite{li2021prefix}), and Adapter Tuning (Adapter \cite{houlsby2019param}, LoRA \cite{hu2022lora}). 

\paragraph{Implementation Details.} To ensure rigorous and fair comparisons, we maintain consistent experimental configurations with UniPT \cite{diao2024unipt} and SHERL \cite{diao2024sherl}, including the optimizer, warm-up scheduler, batch size, training epochs, \emph{etc}. Besides, our method is implemented based on UniPT/SHERL, denoted as Ours${}^{\dagger/\ddagger}$, respectively.
Additional training details are depicted in \textit{Extended Version}.

\subsection{Main Results}
\paragraph{Baselines.} Similar to \cite{diao2024sherl}, in order to conduct a more exhaustive and challenging evaluation, we present a comparison on diverse VL and NLP tasks with various pre-trained architectures, including:
\begin{itemize}
    \item ITR task: \textit{VSE$\infty$} \cite{chen2021VSE} leverages BERT-base as text and Instagram (WSL) pre-trained ResNeXt-101(32×8d) as vision backbones. 
    \item VTR task: \textit{CLIP4Clip} \cite{luo2021clip4clip} adapts pre-trained CLIP's dual-Transformer framework (ViT-B/32 + Text Transformer) through temporal domain adaptation from image-text to video-text spaces.
    \item QA task: \textit{CLIP-ViL} \cite{shen2021much} utilizes frozen CLIP image encoder with text embeddings, followed by a cross-modal fusion Transformer.
    \item VG task: \textit{MDETR} \cite{kamath2021mdetr} combines ResNet-101 and RoBERTa-B for image and text encoding, with a query-attended encoder-decoder Transformer.
    \item NLP task: \textit{T5-series} \cite{raffel2020exploring} imports text encoder and autogressive decoder, with balanced layer reduction (6/24 total layers of side network, equally split for encoder and decoder for \textit{base/large}).
\end{itemize}

{\begin{table*}[!t]
    \centering
    \setlength{\tabcolsep}{5.2pt}
    \begin{center}
    \begin{tabular}{lcccccccccccc}
         \toprule[1.2pt]
         \multirow{2}{*}{\centering Method} & Params. & \multicolumn{2}{c}{Memory ( G) $\downarrow$} & \multirow{2}{*}{CoLA} & \multirow{2}{*}{SST-2} & \multirow{2}{*}{MRPC} & \multirow{2}{*}{QQP} & \multirow{2}{*}{MNLI} & \multirow{2}{*}{QNLI} & \multirow{2}{*}{RTE} & \multirow{2}{*}{STS-B} & \multirow{2}{*}{Avg.} \\
         & (\%) $\downarrow$ & Train & Test& & & & & & & & & \\
         \midrule[0.6pt]
         {{Fully-FT}} &  {100} &  {17.6} &  {0.86} &  {62.8} &  {93.9} &  {91.9} &  {89.9} &  {86.2} &  {92.5} &  {74.1} &  {90.3} &  {85.2} \\
         Adapter & 1.63 & 13.0 & 0.87 & 64.4 & 94.2 & 88.9 & 88.9 & 86.4 & 93.1 & 75.1 & 91.1 & 85.3 \\
         LoRA & 1.71 & 12.6 & 0.86 & 63.3 & 94.3 & 90.1 & 89.0 & 86.3 & 93.2 & 75.5 & 90.9 & 85.3 \\
         BitFit & 0.13 & 10.7 & 0.86 & 61.8 & 94.3 & 91.0 & 88.7 & 85.6 & 93.1 & 67.6 & 90.8 & 84.1 \\
         Prompt & 0.03 & 22.2 & 0.87 & 0 & 90.3 & 74.6 & 88.5 & 82.5 & 92.5 & 59.5 & 90.1 & 72.2 \\
         \midrule[0.6pt]
         LST & 1.74 & 5.5 & 0.88 & 58.1 & 94.1 & 90.4 & 88.8 & \underline{85.6} & 93.3 & \underline{71.9} & 90.7 & 84.1 \\
         UniPT & \underline{1.36} & \textbf{2.9} & \textbf{0.86} & \underline{62.2} & \underline{94.2} & \underline{90.8} & 88.9 & 85.5 & 93.3 & 69.8 & 89.7 & 84.3 \\
         SHERL & \textbf{0.85} & \textbf{2.9} & \underline{0.87} & 61.1 & 93.7 & 89.4 & 88.8 & 85.3 & 93.3 & \underline{71.9} & \underline{90.9} & 84.3 \\
         \textbf{Ours${}^\dagger$} & 2.41 & \underline{3.2} & \textbf{0.86} & \textbf{63.4} & \textbf{94.6} & \textbf{91.6} & \underline{89.1} & 85.7 & \underline{93.4} & 70.9 & 90.0 & \underline{84.8} \\
         \textbf{Ours${}^\ddagger$} & {1.48} & 3.3 & {0.87} & 62.4 & 94.3 & 89.6 & \textbf{89.3} & \textbf{85.9} & \textbf{93.2} & \textbf{72.8} & \textbf{91.5} & \textbf{84.9} \\
         \midrule[0.6pt]
         LST~(T5-large) & 1.23 & 12.2 & 2.88 & 65.3 & 95.7 & 91.6 & 89.7 & {88.6} & 94.1 & 79.9 & \underline{92.4} & 87.1 \\
         UniPT~(T5-large) & 0.92 & 9.1 & \underline{2.82} & 65.7 & \underline{95.8} & 92.0 & 89.7 & 88.2 & 94.2 & 79.6 & 92.0 & 87.2 \\
         SHERL~(T5-large) & \textbf{0.64} & \textbf{7.1} & \textbf{2.80} & 65.6 & \underline{95.8} & {92.9} & 89.6 & {88.6} & 94.2 & \underline{80.8} & 92.1 & 87.5 \\
         \textbf{Ours${}^\dagger$~(T5-large)} & 1.75 & 9.9 & \underline{2.82} & \textbf{66.7} & \textbf{96.5} & \underline{93.1} & \underline{90.0} & \underline{88.7} & \underline{94.7} & 79.8 & \underline{92.4} & \underline{87.7} \\
         \textbf{Ours${}^\ddagger$~(T5-large)} & \underline{0.81} & \underline{7.6} & \textbf{2.80} & \underline{66.4} & \textbf{96.5} & \textbf{93.4} & \textbf{90.3} & \textbf{88.9} & \textbf{94.9} & \textbf{81.6} & \textbf{92.7} & \textbf{88.1} \\
         \bottomrule[1.2pt]
    \end{tabular}
    \end{center}
    \caption{Comparison results with PETL \textbf{(Top)} and METL \textbf{(Bottom)} methods on GLUE benchmark, with \textit{T5-base/large}. We report the number of learnable parameters and memory usage as efficiency metrics, and accuracy, F1 score, Matthew’s Correlation, and Pearson-Spearman Correlation as performance indicators. The best results are highlighted in \textbf{bold}, and the second best results are \underline{underlined}.}     \label{GLUE_table}
\end{table*}}

\paragraph{MP-ISMoE outweights METL methods in memory-constrained scenarios.} 
We compare MP-ISMoE with state-of-the-art METL methods on five VL tasks.
As shown in Table~\ref{VL_table}, our MP-ISMoE achieves superior performance, with the minimal discrepancy from the fully fine-tuned model. Concretely, it demonstrates the following advantages: 
{(1) Remarkable performance improvement.} MP-ISMoE outperforms LST across a range of tasks and backbones. When integrated with UniPT/SHERL (\emph{i.e.}, Ours${}^\dagger$/${}^\ddagger$), it yields an average improvement of 1.4/1.2\% in R@1 and 4.6/4.4\% in Rsum for cross-modal retrieval, 0.80/0.79\% improvements for question answering, and 5.47/4.92\% for visual grounding. These results in challenging pattern matching and limited data-driven scenarios strongly underscore the efficacy of MP-ISMoE. 
{(2) Comparable training memory consumption.} In most cases, MP-ISMoE reduces training memory usage by approximately 50\% compared to LST. Despite a slight memory increase (\emph{e.g.} 1GB for the retrieval task) over the baselines, we consider it acceptable in light of the gains in performance. 
{(3) Negligible inference cost.} With the sparse MoE-based side network where only a fixed number of experts are activated, MP-ISMoE introduces no extra inference cost, which ensures its practical applicability for real-world deployment.

To further assess the generalization ability of our approach, we conduct additional evaluations on NLP tasks.
As shown in Table~\ref{GLUE_table}, MP-ISMoE improves the overall performance of baseline UniPT and SHERL by 0.6\%, with comparable memory consumption. More significantly, it outperforms LST comparable trainable parameters while reducing training memory usage by over 30\%.

In summary, MP-ISMoE enables a larger trainable parameter space and scales up model capacity without notably increasing memory overhead, thereby achieving a more favorable trade-off between memory efficiency and performance.

\begin{table*}[!ht]
    \centering
    {\begin{tabular}{cccc ccc ccc ccc}
         \toprule[1.2pt]
         \multirow{2}{*}{GNP-IQ} & \multirow{2}{*}{ISMoE} & \multirow{2}{*}{\makecell{Params.\\(M)$\downarrow$}} & \multirow{2}{*}{\makecell{Memory\\(G)$\downarrow$}} & \multicolumn{3}{c}{Flickr30K} & \multicolumn{3}{c}{MSCOCO1K} & \multicolumn{3}{c}{MSCOCO5K} \\
         \cmidrule[0.4pt]{5-13}
          & & & & {I-T} $\uparrow$ & {T-I} $\uparrow$ & {Rsum} $\uparrow$ & {I-T} $\uparrow$ & {T-I} $\uparrow$ & {Rsum} $\uparrow$ & {I-T} $\uparrow$ & {T-I} $\uparrow$ & {Rsum} $\uparrow$ \\
         \midrule[0.6pt]
          &  & 12.4 & \underline{24.4} & 84.8 & 69.1 & 537.4 & 80.6 & 67.5 & 532.9 & 61.1 & 45.9 & 445.3 \\
         \checkmark &  & \underline{12.5} & \textbf{18.4} & {83.7} & {68.1} & {534.8} & {78.9} & {66.3} & {530.4} & 59.9 & 44.7 & 442.7 \\
          & \checkmark & {12.7} & {32.4} & \textbf{86.9} & \textbf{71.7} & \textbf{544.3} & \textbf{82.5} & \textbf{69.8} & \textbf{540.4} & \textbf{63.4} & \textbf{47.7} & \textbf{450.9} \\
         \checkmark & \checkmark & 12.9 & {25.5} & \underline{86.5} & \underline{71.3} & \underline{543.1} & \underline{81.9} & \underline{69.1} & \underline{538.7} & \underline{62.2} & \underline{47.1} & \underline{449.1} \\
         \bottomrule[1.2pt]
    \end{tabular}}
    \caption{Ablation results (\%) of the main components using \textit{VSE$\infty$} on ITR tasks. The best results are highlighted in \textbf{bold}, and the second best results are \underline{underlined}.}    \label{main_ablation_table}
\end{table*}

\begin{table*}[!ht]
    \centering
    \resizebox{\textwidth}{!}
    {\begin{tabular}{cccc ccc ccc ccc}
         \toprule[1.2pt]
         \multirow{2}{*}{\makecell{Weight\\Precision}} & \multirow{2}{*}{\makecell{Gaussian \\Noise}} & \multirow{2}{*}{\makecell{Params.\\(M)$\downarrow$}} & \multirow{2}{*}{\makecell{Memory\\(G)$\downarrow$}} & \multicolumn{3}{c}{Flickr30K} & \multicolumn{3}{c}{MSCOCO1K} & \multicolumn{3}{c}{MSCOCO5K} \\
         \cmidrule[0.4pt]{5-13}
          & & & & {I-T} $\uparrow$ & {T-I} $\uparrow$ & {Rsum} $\uparrow$ & {I-T} $\uparrow$ & {T-I} $\uparrow$ & {Rsum} $\uparrow$ & {I-T} $\uparrow$ & {T-I} $\uparrow$ & {Rsum} $\uparrow$ \\
         \midrule[0.6pt]
         Full-Pre &  & \textbf{12.4} & {24.4} & \textbf{84.8} & \textbf{69.1} & \textbf{537.4} & \textbf{80.6} & \textbf{67.5} & \textbf{532.9} & \textbf{61.1} & \textbf{45.9} & \textbf{445.3} \\
         Low-Pre &  & \textbf{12.4} & \textbf{14.8} & 81.6 & 66.4 & 529.3 & 77.5 & 64.6 & 524.7 & 58.4 & 43.0 & 433.8 \\
         Mixed-Pre &  & \textbf{12.4} & \underline{15.4} & 82.1 & 66.7 & 531.4 & 77.8 & 65.0 & 526.9 & 58.8 & 43.5 & 437.2 \\
         Mixed-Pre & \checkmark & \underline{12.5} & {18.4} & \underline{83.7} & \underline{68.1} & \underline{534.8} & \underline{78.9} & \underline{66.3} & \underline{530.4} & \underline{59.9} & \underline{44.7} & \underline{442.7} \\
         \bottomrule[1.2pt]
    \end{tabular}}
    \resizebox{\textwidth}{!}
    {\begin{tabular}{cccc ccc ccc ccc}
         \multirow{2}{*}{\makecell{MoE\\Structure}} & \multirow{2}{*}{\makecell{Network\\Correlation}} & \multirow{2}{*}{\makecell{Params.\\(M)$\downarrow$}} & \multirow{2}{*}{\makecell{Memory\\(G)$\downarrow$}} & \multicolumn{3}{c}{Flickr30K} & \multicolumn{3}{c}{MSCOCO1K} & \multicolumn{3}{c}{MSCOCO1K} \\
         \cmidrule[0.4pt]{5-13}
          & & & & {I-T} $\uparrow$ & {T-I} $\uparrow$ & {Rsum} $\uparrow$ & {I-T} $\uparrow$ & {T-I} $\uparrow$ & {Rsum} $\uparrow$ & {I-T} $\uparrow$ & {T-I} $\uparrow$ & {Rsum} $\uparrow$ \\
         \midrule[0.6pt]
          &  & \textbf{12.4} & \textbf{24.4} & 84.8 & 69.1 & 537.4 & 80.6 & 67.5 & 532.9 & 61.1 & 45.9 & 445.3 \\
          \checkmark &  & \underline{12.7} & \underline{31.6} & \underline{86.3} & \underline{71.3} & \underline{543.1} & \underline{82.0} & \underline{69.2} & \underline{538.8} & \underline{62.9} & \underline{47.1} & \underline{449.4} \\
          \checkmark & \checkmark & \underline{12.7} & {32.4} & \textbf{86.9} & \textbf{71.7} & \textbf{544.3} & \textbf{82.5} & \textbf{69.8} & \textbf{540.4} & \textbf{63.4} & \textbf{47.7} & \textbf{450.9} \\
         \bottomrule[1.2pt]
    \end{tabular}}
    \caption{(\textbf{Top}) Ablation results (\%) of GNP-IQ with various backbone weights precision (Full-, Mixed-, and Low-Precision), w/ or w/o Gaussian noise perturbation. (\textbf{Bottom}) Ablation results (\%) of ISMoE w/ or w/o MoE structure and measuring correlation between networks. All experiments are conducted on ITR tasks with \textit{VSE$\infty$}. The best results are highlighted in \textbf{bold}, and the second best results are \underline{underlined}.}     \label{GNP-IQ_ISMoE_ablation_table}
\end{table*}

\paragraph{MP-ISMoE outperforms PETL methods with similar training memory consumption.} 
We also evaluate the proposed method with state-of-the-art PETL methods on the GLUE benchmark for NLP tasks.
As shown in Table~\ref{GLUE_table}, with the \textit{T5-base} model as backbone, MP-ISMoE significantly reduces the training memory overhead from 17.6GB to 3.2GB, up to 81.8\% of the fully fine-tuning, while in context of the same scale of trainable parameters, the prevailing Adapter and LoRA methods only gain a reduction ratio of 25.6\%. 
Besides, MP-ISMoE surpasses BitFit and Prompt by 0.8\% and 12.7\% on average, respectively, while requiring only 29.9\% and 14.4\% of their training memory overhead.
These results demonstrate that MP-ISMoE achieves significantly higher memory efficiency than both Full-FT and other PETL methods.
To further exploit the memory efficiency and validate the scalability on larger backbones, we continue our comparison on the \textit{T5-large} backbone. 
Remarkably, MP-ISMoE achieves a 15.7\% performance gain over Prompt under similar or even lower training memory consumption, and consistently outperforms other PETL baselines without incurring additional inference memory overhead.

\subsection{Ablation Study}
\paragraph{On Main Components.}
We evaluate the effect of the main components, including Gaussian Noise Perturbed Iterative Quantization (GNP-IQ) and Interactive Side Mix-of-Experts (ISMoE), on the \textit{VSE$\infty$} for ITR task.
Here, we use UniPT as baseline.
As summarized in Table~\ref{main_ablation_table}, GNP-IQ significantly reduces training memory consumption from 24.4GB to 18.4GB (a reduction of 24.6\%) by quantizing the pre-trained backbone into lower-bit precision weights. Although this inevitably leads to a slight performance degradation, it frees up substantial memory for scaling up the side network.
Conversely, the single introduction of ISMoE significantly boosts performance, by improving R@1 and Rsum by 2.2\% and 6.7\%, respectively, at a cost of increased training memory usage.
These results highlight the inherent strengths of the two modules, \emph{i.e.} the memory efficiency of GNP-IQ and the accuracy advantage of ISMoE.
When further combined, GNP-IQ and ISMoE complement each other by reallocating part of the memory budget from the backbone to the expanded side network, ultimately yielding average improvements of 1.5\% in R@1 and 5.1\% in RSum, respectively, with only a negligible increase in memory overhead.

\paragraph{On Effect of GNP-IQ.} 
We further evaluate the individual effect of designs in GNP-IQ on the \textit{VSE$\infty$} \cite{chen2021VSE} for ITR task.
As shown in the Table~\ref{GNP-IQ_ISMoE_ablation_table}, on the basis of the baseline, we first fine-tune with the frozen pre-trained backbone using different weight precision formats.
Specifically, the introductions of low unified precision and mixed-precision reduce training memory by 36.9\% and 39.3\%, respectively.
However, the retrieval accuracy incurs severe degradation, although the drop under mixed-precision fine-tuning is relatively moderate.
Furthermore, under mixed-precision, the introduction of Gaussian noise perturbation enables recovery of retrieval accuracy while maintaining a relatively low memory consumption. 
This is attributed to the noise-induced perturbations effectively simulating long-term weight updates, thereby mitigating the accumulated quantization error that would otherwise impair fine-tuning.

\paragraph{On Effect of ISMoE.} 
We also evaluate the effect of detailed designs in ISMoE on the \textit{VSE$\infty$} \cite{chen2021VSE} for ITR task.
As previously discussed, the memory overhead introduced by this module can be compensated by the reduction achieved from the GNP-IQ module, therefore, memory consumption is not the focus of this section.
As shown in Table~\ref{GNP-IQ_ISMoE_ablation_table}, introducing the sparse MoE structure significantly improves the R@1 and Rsum by 1.6\% and 5.2\%, respectively. This result underscores the effectiveness of MoE-based scaling up in enhancing transfer learning.
Upon further incorporating expert selection based on salient token from the backbone, these metrics are continuously increased by 2.2\% and 6.7\%, respectively, while keeping the amount of learnable parameters and memory consumption basically constant.
This improvement can be attributed to the more effective utilization of general knowledge from the backbone in guiding expert selection, which in turn alleviates over-fitting and mitigates the forgetting of general knowledge.

We also extensively study the \textbf{influence of the ratio $\bm{p}$ of re-quantized weights} in each iteration, \textbf{impact of the number of experts $\bm{N}$} in Eq.~\eqref{moe_gate} in MoE structures. Due to space limitation, we summarize the detailed results in \emph{Extended Version}.

\section{Conclusion}
In this paper, we propose a novel METL method dubbed Mixed-Precision Interactive Side Mixture-of-Experts (MP-ISMoE), which effectively addresses the inherent limitations of existing methods regarding the scalability and representational capabilities of side networks.
We develop the Gaussian Noise Perturbed Iterative Quantization (GNP-IQ) process that enables mixed-precision training, effectively compressing the memory footprint of the backbone while preserving more performance. Furthermore, the Interactive Side Mixture-of-Experts (ISMoE) structure is introduced, scaling up the side network by reallocating the previously saved memory, and mitigating knowledge forgetting by leveraging salient token-guided expert selection.
Experimental results on multiple vision-language and natural language processing tasks demonstrate that our method achieves superior balance between trainable parameters, memory efficiency and transfer learning performance, by surpassing existing state-of-the-art METL methods in accuracy with comparable memory overhead.

\section{Acknowledgements}
This work was partly by the National Natural Science Foundation of China (No. 62202034), the Beijing Natural Science Foundation (No. 4242044), the Aeronautical Science Foundation of China (No. 2023Z071051002), CCF Baidu Open Fund, the Graduate Education and Development Research Special Fund of Beihang University, and the Fundamental Research Funds for the Central Universities.

\bibliography{main}

\end{document}


\maketitle



{\color{red}In this document, we additionally provide detailed introduction of datasets and metrics in Sec.A, report more implementation details of the experiments in Sec.B, and present more ablation studies on certain hyper-parameters }

\section{A. Datasets and Metrics}
\label{SecA}
\textbf{Image-Text Retrieval (ITR)} employs Flickr30k \cite{young2014image} and MSCOCO \cite{lin2014microsoft} for image-text matching tasks. Flickr30k comprises 31,783 colloquial images with 158,915 captions distributed across 29,783 training, 1,000 validation, and 1,000 test instances, where captions exhibit explicit compositional semantics emphasizing object-attribute-spatial relationships that necessitate fine-grained matching capabilities. MSCOCO contains 123,287 complex scene images annotated with 616,435 textual descriptions, utilizing the standard Karpathy split of 113,287 training, 5,000 validation, and 5,000 test images; its intricate contextual object interactions and scene dynamics demand deeper relational reasoning. Both benchmarks evaluate bidirectional retrieval performance through Image-to-Text (I-T) and Text-to-Image (T-I) Recall@1, while RSum—the summation of Recall@K scores at K=1,5,10 for both retrieval directions—serves as the holistic metric capturing compositional complexity.

\textbf{Video-Text Retrieval (VTR)} utilizes MSR-VTT \cite{xu2016msr} and MSVD \cite{chen2011collecting} for cross-modal retrieval. MSR-VTT (Microsoft Research Video to Text) comprises 10,000 YouTube-sourced video clips aggregating ~41 hours of content, each annotated with 20 English captions. Following the standard 1k-A protocol, we partition the dataset into 9,000 training videos with full captions and 1,000 test pairs. Characterized by high diversity across 20 categories (e.g., sports, music), its captions describe complex temporal dynamics and object interactions. MSVD (Microsoft Video Description) contains 1,970 short video clips with approximately 80,000 multilingual annotations, divided into 1,200 training, 100 validation, and 670 test videos. Noted for fine-grained temporal alignment challenges, MSVD emphasizes precise action-object localization. Both benchmarks evaluate bidirectional retrieval performance through Video-to-Text (V-T) and Text-to-Video (T-V) Recall@1 and RSum metrics.

\textbf{Question Answering (VQA\&GQA)} employs VQAv2 \cite{goyal2017making} and GQA \cite{hudson2019gqa} for visual reasoning tasks. VQAv2 mitigates language bias through 204,721 COCO images paired with 1,105,904 questions, utilizing standard splits: 82,783 training images (443,757 questions), 40,504 validation images (214,354 questions), and 81,434 test images (447,793 questions). Its questions necessitate diverse reasoning about object attributes, actions, and scene context. GQA features 113,018 images with 22,669,678 scene graph-generated questions ensuring compositional rigor, distributed across balanced partitions: approximately 70

\textbf{Visual Grounding (VG)} employs RefCOCO, RefCOCO+ \cite{yu2016modeling}, and RefCOCOg \cite{mao2016generation} derived from MSCOCO for reference comprehension. RefCOCO contains 19,994 images with 50,000 bounding boxes annotated by 142,210 expressions, utilizing the UNC split: 120,624 training expressions, 10,834 validation expressions, and Test A/B partitions (5,675 person-centric/5,095 non-person instances). RefCOCO+ shares the image corpus but imposes stricter constraints—prohibiting location words in its 141,564 expressions describing 49,856 objects—with similar splits (120,191 train, 10,758 val, 5,726/4,889 Test A/B). RefCOCOg substantially differs with 26,711 images, 54,822 objects, and 104,560 grammatically complex expressions, partitioned into 85,474 training, 7,323 validation, and 9,592 test samples. Evaluation measures localization accuracy via Precision@0.5 (IoU threshold 0.5) between predicted and human-annotated bounding boxes.

{\begin{table*}[!ht]
    \centering
    \vskip 0.1in
    \begin{center}
    {
    \begin{tabular}{lcccccc}
         \toprule[1.2pt]
         Task & Model & Learning Rate & Optimizer ($\beta_1$, $\beta_2$, Weight Decay) & Batch Size & Total Epochs & Warmup Strategy \\
         \midrule[0.6pt]
         ITR & VSE$\infty$ & $5 \times 10^{-4}$ & $0.9, 0.999, 1 \times 10^{-2}$ & 112 & 25 & linear \\
         VTR & CLIP4Clip & $1 \times 10^{-4}$ & $0.9, 0.98, 1 \times 10^{-2}$ & 128 & 5 & cosine \\
         VQA & CLIP-ViL & $5 \times 10^{-4}$ & $0.9, 0.999, 1 \times 10^{-2}$ & 256 & 5 & linear \\
         GQA & CLIP-ViL & $1 \times 10^{-4}$ & $0.9, 0.999, 1 \times 10^{-2}$ & 256 & 5 & linear \\
         VG & MDETR & $5 \times 10^{-4}$ & $0.9, 0.999, 0$ & 8 & 10 & linear \\
         \bottomrule[1.2pt]
    \end{tabular}
    }
    \end{center}
    \caption*{Table A: Detailed Hyper-parameters of MDPD on ITR, VTR, VQA, GQA, and VG tasks. Among them, \textit{AdamW} is adopted as the optimizer uniformly.}  \label{tab_vl_hyperparameters}
\end{table*}}

\begin{table*}[!t]
    \centering
    {\begin{tabular}{ccc ccc ccc ccc}
         \toprule[1.2pt]
         \multirow{2}{*}{$p$ \#} & {Params.} & {Memory} & \multicolumn{3}{c}{Flickr30K} & \multicolumn{3}{c}{MSCOCO1K} & \multicolumn{3}{c}{MSCOCO5K} \\
         \cmidrule[0.4pt]{4-12}
         & (M) $\downarrow$ & (G) $\downarrow$ & {I-T} $\uparrow$ & {T-I} $\uparrow$ & {Rsum} $\uparrow$ & {I-T} $\uparrow$ & {T-I} $\uparrow$ & {Rsum} $\uparrow$ & {I-T} $\uparrow$ & {T-I} $\uparrow$ & {Rsum} $\uparrow$ \\
         \midrule[0.6pt]
         0\% & \textbf{12.8} & \textbf{22.5} & 84.9 & 69.9 & 539.7 & 80.8 & 67.8 & 535.2 & 61.1 & 45.9 & 444.6\\
         5\% & \underline{12.9} & \underline{24.2} & 86.0 & 70.7 & 541.9 & 81.5 & \underline{68.6} & 537.1 & 61.7 & 46.9 & 447.4\\
         10\% & \underline{12.9} & 25.5 & \textbf{86.5} & \underline{71.3} & \underline{543.1} & \underline{81.9} & \textbf{69.1} & \textbf{538.7} & \underline{62.2} & \underline{47.1} & \underline{449.1}\\
         50\% & \underline{12.9} & 30.4 & \underline{86.3} & \textbf{71.5} & \textbf{543.4} & \textbf{82.1} & \underline{68.6} & \underline{538.4} & \textbf{62.7} & \textbf{47.2} & \textbf{449.4}\\
         \bottomrule[1.2pt]
    \end{tabular}}
\caption*{Table B: Ablation results (\%) on the ratio $p$ of re-quantized weights on distinct datasets for ITR task. The best results are highlighted in \textbf{bold}. and the second best results are \underline{underlined}.}    \label{tab_ablation_p}
\end{table*}

\begin{table*}[!t]
    \centering
    {\begin{tabular}{ccc ccc ccc ccc}
         \toprule[1.2pt]
         \multirow{2}{*}{Experts \#} & {Params.} & {Memory} & \multicolumn{3}{c}{Flickr30K} & \multicolumn{3}{c}{MSCOCO1K} & \multicolumn{3}{c}{MSCOCO5K} \\
         \cmidrule[0.4pt]{4-12}
         & (M) $\downarrow$ & (G) $\downarrow$ & {I-T} $\uparrow$ & {T-I} $\uparrow$ & {Rsum} $\uparrow$ & {I-T} $\uparrow$ & {T-I} $\uparrow$ & {Rsum} $\uparrow$ & {I-T} $\uparrow$ & {T-I} $\uparrow$ & {Rsum} $\uparrow$ \\
         \midrule[0.6pt]
         3 & \textbf{12.7} & \textbf{25.0} & 85.7 & 70.1 & 539.6 & 81.2 & 68.0 & 535.7 & 61.5 & 46.2 & 446.7 \\ 
         4 & \underline{12.8} & \underline{25.2} & 86.0 & 70.8 & 541.2 & 81.5 & 68.6 & 537.6 & 61.4 & 46.4 & 447.3 \\ 
         5 & 12.7 & 25.3 & 86.2 & 70.9 & 541.9 & 81.4 & 68.9 & 538.1 & 61.8 & 46.9 & 448.2 \\ 
         6 & 12.9 & 25.5 & \underline{86.5} & 71.3 & 543.1 & 81.9 & 69.1 & 538.7 & 62.2 & 47.1 & 449.1 \\ 
         7 & 13.1 & 25.8 & 86.3 & \textbf{71.8} & \underline{543.7} & \underline{82.3} & \underline{69.6} & \underline{539.2} & \textbf{62.7} & \underline{47.5} & \underline{450.3} \\ 
         8 & 13.2 & 26.0 & \textbf{86.9} & \underline{71.6} & \textbf{544.0} & \textbf{82.6} & \textbf{70.1} & \textbf{539.8} & \underline{62.5} & \textbf{47.9} & \textbf{450.5} \\ 
         \bottomrule[1.2pt]
    \end{tabular}}
\caption*{Table C: Ablation results (\%) on the total number of experts in the side network on distinct datasets for ITR task. The best results are highlighted in \textbf{bold}. and the second best results are \underline{underlined}.}    \label{tab_ablation_N}
\end{table*}

\textbf{Language-only task} utilizez the General Language Understanding Evaluation (GLUE) benchmark \cite{wang2018glue}, which integrates eight natural language processing tasks across four core categories: linguistic acceptability evaluated via CoLA \cite{warstadt2019neural}, sentiment analysis using SST-2 \cite{socher2013recursive}, similarity and paraphrase detection encompassing MRPC \cite{dolan2005automatically}, QQP, and STS-B \cite{cer2017semeval}, along with natural language inference covering MNLI \cite{williams2017broad}, QNLI \cite{rajpurkar2016squad}, and RTE \cite{bentivogli2009fifth}. Task-specific evaluation metrics include: Accuracy for classification tasks (SST-2, MNLI, RTE, QNLI); F1-score supplemented by Accuracy for paraphrase detection (MRPC, QQP); Matthew's Correlation Coefficient addressing class imbalance in CoLA; and Pearson-Spearman Correlation for semantic similarity assessment in STS-B.

\section{B. Implementation Details}
\label{SecB}
For vision-language (VL) tasks, the hyper-parameter configurations are detailed in Table \ref{tab_vl_hyperparameters}. Specifically, on the ITR task using VSE$\infty$, we employ a batch size of 112 and maintain consistency with pre-trained models for all other tasks, the learning rate and reduction factor $r$ to $10\times lr$ and 2 respectively and weight-only quantization is performed iteratively at task-specific intervals, \emph{i.e.}, every $M=10/2/2/4$ epochs for ITR/VTR/QA/VG tasks. For NLP tasks, the learning rate and $r$ are adjusted to $3\times10^{-3}$ and 8, while each method undergoes training with 20 epochs on small datasets and 10 epochs on larger ones, and re-quantization occurs every $M=8/4$ epochs for small/large datasets. Following LST~\cite{sung2022lst}, we implement a layer-dropping strategy: for \textit{T5-base}, this removes the $0^{\text{th}}$, $4^{\text{th}}$, and $8^{\text{th}}$ layers in both the encoder and decoder; for \textit{T5-large}, all even-indexed encoder and decoder layers are omitted. Besides, during training, $p=10\%$ of frozen backbone weights are randomly selected for quantization per time, the number of experts extended $N$ and routed $K$ in the side network is 6 and 1 respectively, and the weight of supervised fine-tuning loss $\alpha$ and load balancing loss $\beta$ is set to 1 and $1\times10^{-3}$. All experiments are executed on two NVIDIA GeForce RTX 3090Ti GPUs.

\section{C. Backpropagation through Large Backbone}
Consider a neural network comprising $L$ sequential layers, where the transformation at the $i^{\text{th}}$ layer is defined as $f_i(\mathbf{x}) = \sigma_i(\mathbf{W}_i \mathbf{x} + \mathbf{b}_i)$. This composite function depends on the previous layer's output, parameterized by the weight matrix $\mathbf{W}_i$, bias vector $\mathbf{b}_i$, and nonlinear activation function $\sigma_i(\cdot)$. We denote the pre-activation output as $\mathbf{z}_{i+1}$ and the post-activation output as $\mathbf{a}_{i+1}$, establishing the layer-wise propagation:

\begin{equation}
    \mathbf{a}_{i+1} = \sigma_i(\mathbf{z}_{i+1}) = \sigma_i(\mathbf{W}_i \mathbf{a}_i + \mathbf{b}_i).
\end{equation}

Network parameters are optimized via stochastic gradient descent (SGD) by minimizing a scalar loss function $\mathcal{L}$ applied to the final layer output. The backpropagation algorithm computes gradients for $\mathbf{W}_i$ and $\mathbf{b}_i$ through recursive application of the multivariate chain rule:

\begin{equation}
\label{wb_bp}
    \begin{aligned}
        \frac{\partial \mathcal{L}}{\partial \mathbf{W}_i} 
        &= \frac{\partial \mathcal{L}}{\partial \mathbf{a}_{i+1}} 
           \frac{\partial \mathbf{a}_{i+1}}{\partial \mathbf{z}_{i+1}}
           \frac{\partial \mathbf{z}_{i+1}}{\partial \mathbf{W}_i} 
        = \frac{\partial \mathcal{L}}{\partial \mathbf{a}_{i+1}} \sigma_i' \mathbf{a}_i^\top, \\
        \frac{\partial \mathcal{L}}{\partial \mathbf{b}_i} 
        &= \frac{\partial \mathcal{L}}{\partial \mathbf{a}_{i+1}} 
           \frac{\partial \mathbf{a}_{i+1}}{\partial \mathbf{z}_{i+1}} 
        = \frac{\partial \mathcal{L}}{\partial \mathbf{a}_{i+1}} \sigma_i',
    \end{aligned}
\end{equation}
where $\sigma_i' \equiv \frac{d\sigma_i}{d\mathbf{z}_{i+1}}$ denotes the activation gradient, and $\frac{\partial \mathcal{L}}{\partial \mathbf{a}_{i+1}}$ represents the upstream gradient from subsequent layers. This upstream gradient is recursively computed via backward propagation from layer $i+2$:

\begin{equation}
\label{a_bp}
    \frac{\partial \mathcal{L}}{\partial \mathbf{a}_{i+1}} 
    = \frac{\partial \mathcal{L}}{\partial \mathbf{a}_{i+2}} 
      \frac{\partial \mathbf{a}_{i+2}}{\partial \mathbf{z}_{i+2}} 
      \frac{\partial \mathbf{z}_{i+2}}{\partial \mathbf{a}_{i+1}} 
    = \frac{\partial \mathcal{L}}{\partial \mathbf{a}_{i+2}} \sigma_{i+1}' \mathbf{W}_{i+1}^\top.
\end{equation}

As formalized in Equations \eqref{wb_bp} and \eqref{a_bp}, the backpropagation algorithm incurs substantial computational overhead due to floating-point operations (FLOPs) required for two critical gradient components: 1) The activation gradients $\{\mathbf{a}\}$ corresponding to updated parameters $\{\mathbf{W}\}$, and 2) The activation derivatives $\{\sigma'\}$ that must be cached throughout the computational graph, where $\{\cdot\}$ denotes sets of activations, parameters, or gradients. Existing Parameter-Efficient Transfer Learning (PETL) techniques, including Adapter~\cite{houlsby2019param}, Prompt-Tuning~\cite{lester2021power}, and LoRA~\cite{hu2022lora}, mitigate memory footprint by reducing the parameter update set $|\{\mathbf{W}\}|$ through learning only sparse parameter subsets, where $|\{\cdot\}|$ means the size of set $\{\cdot\}$. Consequently, the memory allocated for activation storage $|\{\mathbf{a}\}|$ proportionally decreases. However, the dominant computational burden during backpropagation stems from computing gradient terms involving $\{\sigma'\}$ - the derivatives of activation functions. Crucially, $|\{\sigma'\}|$ remains undiminished in these methods since:
\begin{equation}
|\{\sigma'\}| = \sum_{i=1}^{L} \dim(\mathbf{z}_i).
\end{equation}

This persistence occurs because PETL methods typically introduce trainable parameters into network inputs or intermediate structures while keeping the backbone frozen. Nevertheless, they still require full computation of $\sigma'$ across all backbone operations, necessitating: 1) Complete evaluation of activation gradients through the entire computational graph, 2) Storage of intermediate derivatives at each layer, and 3) Backpropagation through all nonlinear transformations.

Since activation dimensions generally satisfy $|\{\mathbf{a}\}| = |\{\sigma'\}|$ (barring dimensionality-altering activations), the theoretical memory reduction ceiling becomes:
\begin{equation}
\text{Memory}_{\text{BP}} = \underbrace{|\{\mathbf{a}\}|}_{\text{reduced}} + \underbrace{|\{\sigma'\}|}_{\text{unchanged}} \leq 50\% \text{ reduction}.
\end{equation}

Therefore, while PETL methods reduce parameter update costs, they still incur substantial FLOPs and memory requirements proportional to backbone complexity, as full error backpropagation through frozen layers remains mandatory.

Based on the foregoing computational analysis, \textbf{side network} is proposed as a memory-efficient alternative. This lightweight network maintains same structure to the backbone network while scaling all weight matrices and hidden state dimensions by a reduction factor $r \geq 2$. Thus, the original backpropagation memory footprint $|\{\mathbf{a}\}| + |\{\sigma'\}|$ is fundamentally transformed in this paradigm. Crucially, the side network \textit{decouples} from the backbone's computational graph during backpropagation, requiring gradient computation only through its own structure. Consequently, its memory consumption reduces to:
\begin{equation}
\text{Memory}_{\text{BP}}^{\text{side}} = \frac{|\{\mathbf{a}\}| + |\{\sigma'\}|}{r}
\end{equation}

This yields a critical comparative advantage: when $r > 2$, the side network achieves strictly lower memory consumption than the theoretical minimum of Parameter-Efficient Transfer Learning (PETL) methods, which remain bounded by:
\begin{equation}
\text{Memory}_{\text{BP}}^{\text{PETL}} \geq \frac{|\{\mathbf{a}\}| + |\{\sigma'\}|}{2}
\end{equation}

Thus, side networks establish a new efficiency frontier for Memory-Efficient Transfer Learning (METL), with memory savings growing linearly with $r$ while maintaining functional capacity.

\section{D. Baselines}
We select various transfer paradigms for comprehensive and challenging validation:

-\textit{VSE$\infty$} \cite{chen2021VSE} with BERT-base \cite{devllin2019bert} model and ResNeXt-101(32$\times$8d) \cite{xie2017aggregated} backbone pre-trained on Instagram (WSL) on Flickr30K \cite{young2014image}, MSCOCO1K and MSCOCO5K \cite{lin2014microsoft} for the \textbf{ITR} task;

-\textit{CLIP4Clip} \cite{luo2021clip4clip} with the pre-trained CLIP \cite{radford2021learning} using Text Transformer \cite{radford2019language} and ViT-B/32 \cite{alexey2021vit} on MSR-VTT \cite{xu2016msr} and MSVD \cite{chen2011collecting} for the \textbf{VTR} task;

-\textit{CLIP-ViL} \cite{shen2021much} that applies the CLIP image backbone \cite{radford2021learning} and encodes the text into word embeddings, followed by a cross-modal Transformer on VQAv2 \cite{goyal2017making} and GQA \cite{hudson2019gqa} for the \textbf{QA} task;

-\textit{MDETR} \cite{kamath2021mdetr} that integrates a pre-trained ResNet-101, RoBERTa-base \cite{liu2019roberta} with an encoder-decoder Transformer on RefCOCO, RefCOCO+ \cite{yu2016modeling} and RefCOCOg \cite{mao2016generation} for the \textbf{VG} task;

-\textit{T5-series} \cite{raffel2020exploring} that imports text encoder and auto-regressive decoder, while following \cite{sung2022lst}, we drop 6, 24 layers of side network (3, 12 layers each in encoder and decoder) for \textit{T5-base} and \textit{T5-large} on GLUE benchmark \cite{wang2018glue} for the \textbf{NLP} task;

\textit{ViT-base} \cite{alexey2021vit} witch consists of 86 million parameters, while pre-trained on ImageNet-21K \cite{deng2009imagenet} is the most commonly used backbone across prior works (\emph{e.g.}, image classification, video classification, \emph{etc}.), and is adopted on VTAB-1K \cite{zhai2019large} for the \textbf{CV} task.

\section{E. More Ablation Studies}
\label{SecC}
We extensively conduct more ablation studies to verify the effectiveness of the selected hyper-parameters.

\paragraph{Influence of the ratio $p$ of re-quantized weights.}
To further explore the impact of the re-quantization ratio $p$ of frozen backbone weights on model overhead and performance during fine-tuning, we conducted related ablation experiments. As shown in Table~\ref{tab_ablation_p}, re-quantizing only 10\% of the frozen weights achieves optimal performance without incurring heavy training overhead, which means that by using a smaller fraction of weights, better quantization parameters can be obtained, leading to optimal model performance. While re-quantizing less than 10\% of the frozen weights introduces less memory overhead, it is unlikely to significantly improve model performance compared to the baseline. Re-quantizing more weights not only fails to achieve parameter- and memory-efficient training, but can also lead to performance degradation due to over-fitting. Therefore, in GNP-IQ, we randomly select only $p=10\%$ of the frozen weights for req-uantization to update the quantization coefficients.

\paragraph{Impact of the number of experts $N$.}
To explore the influence of the total number of experts in the extended side network and verify the effectiveness of adopting $N=6$, we conduct an ablation study of $N$ on ITR task with \textit{VSE$\infty$}. As demonstrated in Table~\ref{tab_ablation_N}, the results reveal that the performance of our method is sensitive to $N$. Specifically, the model achieves the balance between the amount of activated parameters, training memory footprint, and model performance when $N=6$ on Flickr30K, MSCOCO1K, and MSCOCO5K datasets, indicating that adopting 6 experts to extend the side network's FFN layer ensures that all input features are appropriately and sparsely distributed among all experts. Each expert is able to acquire knowledge belong to their \textit{specific area} of expertise and effectively process each input. Furthermore, this setup maintains a manageable number of trainable parameters and training memory consumption, ensuring boosted model performance while keeping comparable training overhead.
As $N$ increases beyond 6, even if model performance continues to improve, it means greater training overhead, which contradicts the ultimate goal of parameter- and memory-efficient transfer learning. While $N$ is less than 6, model performance does not improve significantly, making it difficult to prove the effectiveness of the method. In summary, in ISMoE, we adopt an MoE with $N=6$ experts to sparsely scale up the parameter size of the side network.

\bibliography{main}